\documentclass{article}
\usepackage{graphicx} 
\usepackage{algorithm} 
\usepackage{algorithmic} 
\usepackage{amsmath} 
\usepackage{hyperref}
\usepackage{booktabs} 
\usepackage{siunitx}
\usepackage[utf8]{inputenc}
\usepackage{authblk}
\usepackage{url}

\title{\textbf{A Unified Module for Accelerating STABLE-DIFFUSION: LCM-LORA}}

\author{Ayush Thakur}
\author{Rashmi Vashisth}

\affil{\small Amity Institute of Information Technology, Amity University Uttar Pradesh, Noida, \texttt{ayush.th2002@gmail.com} \& \texttt{rvashisth@amity.edu}}

\date{Corresponding Author*: Ayush Thakur*}

\begin{document}

\maketitle

\begin{abstract}
    This paper presents a comprehensive study on the unified module for accelerating stable-diffusion processes, specifically focusing on the LCM-LORA module. Stable-diffusion processes play a crucial role in various scientific and engineering domains, and their acceleration is of paramount importance for efficient computational performance. The standard iterative procedures for solving fixed-source discrete-ordinates problems often exhibit slow convergence, particularly in optically thick scenarios. To address this challenge, unconditionally stable diffusion-acceleration methods have been developed, aiming to enhance the computational efficiency of transport equations and discrete ordinates problems. This study delves into the theoretical foundations and numerical results of unconditionally stable diffusion-synthetic acceleration methods, providing insights into their stability and performance for model discrete ordinates problems. Furthermore, the paper explores recent advancements in diffusion model acceleration, including on-device acceleration of large diffusion models via GPU-aware optimizations, highlighting the potential for significantly improved inference latency. The results and analyses in this study provide important insights into stable-diffusion processes and have important ramifications for the creation and application of acceleration methods—specifically, the LCM-LORA module—in a variety of computing environments.
\end{abstract}

\newpage

\section{Introduction}
Latent diffusion models (LDMs) are a type of generative models that can produce highly realistic and diverse images from various inputs, such as natural language or sketches. The concept behind LDMs is to use an autoencoder to convert high-resolution data into a compressed, lower-dimensional latent space, from which a diffusion model may be more effectively trained. Reversing a diffusion process that progressively taints data into pure noise is how diffusion models learn to produce new data \cite{rombach2022high}. Diffusion models applied in the latent space allow LDMs to deliver state-of-the-art picture synthesis outcomes while preserving the intricacies of the data and reducing complexity.

But LDMs also have to deal with certain difficulties, particularly when it comes to the speed and calibre of the picture production process. LDMs use a slower reverse sampling technique that generates a picture from an input by sampling and refining it several times. The real-time application and user experience of LDMs may be restricted by this laborious and computationally costly procedure \cite{yang2022diffusion}. Even with certain open-source models and acceleration methods available, real-time generation on mainstream consumer GPUs remains unattainable for LDMs.

There are two main categories of methods that aim to accelerate LDMs: 
\begin{itemize}
    \item In the first category, reverse sampling is sped up by employing sophisticated ODE solvers, such as DDIM, DPM-Solver, and DPM-Solver++ \cite{andersson2015assimulo, yang2022diffusion}. By utilising adaptive step sizes and interpolation techniques, these approaches can decrease the number of inference steps required to build a picture. But these approaches also have a large computational cost, particularly when they use classifier-free guiding, a method that allows controlling the picture production without requiring the mode to be retrained.
    \item In order to make LDMs more efficient, the second category of approaches include distilling them, which entails compressing and simplifying the model architecture and parameters \cite{meng2023distillation}. Guided-Distill is one example of a distillation technique that may reduce large data sets (LDMs) into more manageable and efficient models while maintaining the flexibility and calibre of the picture production process. Distillation techniques do, however, have several practical drawbacks, such as the high processing costs associated with training the distilled models and the possibility of information and diversity loss.
\end{itemize}

As a result, it's still difficult in practise to balance speed and quality in LDM-generated images. Further research and development are needed to optimise and enhance LDMs and apply them to other activities and domains. Although LDMs are a promising method for picture synthesis, there is still opportunity for improvement and creativity.

\subsection{Latent Consistency Models}
Creating realistic and varied images from a variety of inputs, such text or drawings, is a difficult challenge in image creation. One of the state-of-the-art techniques for image generation is LDMs, which can produce high-resolution images by mapping the data into a lower-dimensional latent space and applying a diffusion model in that space. However, LDMs suffer from a slow sampling issue, as they need to perform multiple steps of sampling and refinement to generate an image from an input. This issue can limit the real-time application and the user experience of LDMs.

To address this issue, Luo proposed latent consistency models (LCMs) \cite{luo2023latent}, which are inspired by consistency models (CMs). CMs are a novel class of generative models that can generate outputs in one step by using a consistency mapping technique, which maps each point along the trajectory of an ordinary differential equation back to its initial point. LCMs extend the idea of CMs to the latent space and the reverse diffusion process, which is the process of generating an output by reversing the diffusion process that gradually corrupts an output into pure noise. The development of the joint distribution of the input and output over the diffusion period is described by the enhanced probability flow ODE (PF-ODE) that LCMs use to design the reverse diffusion process. Then, without the requirement for iterative solutions using numerical ODE-Solvers, LCMs employ the consistency mapping approach to directly anticipate the solution of the PF-ODE in the latent space \cite{song2023consistency}. Taking just one to four inference steps, this leads to an incredibly efficient synthesis of high-resolution images \cite{luo2023lcm}.

LCMs also have an advantage in terms of distillation efficiency, which means the efficiency of simplifying and compressing the model architecture and the parameters to make them more efficient. Pre-trained classifier-free guided diffusion models, or models that can regulate the image generation process without requiring retraining, may be effectively reduced to LCMs. While maintaining the flexibility and quality of the picture production process, LCMs may condense these models into smaller, quicker models. Compared to the previous approaches, LCMs need just 32 A100 GPU hours to train a minimal-step inference model\cite{luo2023lcm}.

These are a potential method for creating images since they can produce high-quality images in fewer stages and get over the slow sampling problem with LDMs. LCMs can also distill pre-trained models with minimal changes and improve the computational efficiency and the storage requirements of image generation. LCMs are a novel and effective way to adapt LDMs and CMs to the latent space and the reverse diffusion process.

\subsection{Latent Consistency Finetuning}
Building on this, LCMs are a novel technique to accelerate the image generation process of LDMs by reducing the number of inference steps. However, to apply LCMs to different image domains, such as anime, photo-realistic, or fantasy images, one needs to fine-tune the pre-trained LCMs using Latent Consistency Finetuning (LCF) \cite{luo2023latent}, which requires access to the teacher diffusion model. Alternatively, one can use Latent Consistency Distillation (LCD) \cite{luo2023latent} to distill a pre-trained LDM into an LCM, or directly fine-tune an LCM using LCF, but these methods are still time-consuming and resource-intensive. Therefore, a natural question arises: is it possible to achieve fast and training-free inference on custom datasets using LCMs?

In this paper, we will go through the LCM-LoRA, a training-free acceleration module for LCMs, which can be easily plugged into various fine-tuned Stable-Diffusion \cite{rombach2022high} models or SD LoRAs \cite{hu2021lora} to enable rapid image generation with minimal steps. LCM-LoRA is based on Low-Rank Adaptation, a performance-efficient fine-tuning technique that adds a small number of adapter layers to the original model. LCM-LoRA leverages LoRA to distill the knowledge of the teacher diffusion model into the adapter layers, which can then be transferred to any fine-tuned SD model or LoRA without additional training. LCM-LoRA can be viewed as a novel type of neural network-based probability flow ODE (PF-ODE) solver, which differs from previous numerical PF-ODE solvers such as DDIM \cite{song2020denoising}, DPM-Solver \cite{lu2022dpm}, and DPM-Solver++ \cite{lu2022dpmpp}. We show that LCM-LoRA has strong generalization abilities across various fine-tuned SD models and LoRAs, and can generate high-quality images with few-step inference.

\section{Literature Survey}
LCM-LoRA is based on LCM, which are a recent innovation that reduces the number of inference steps required by LDMs by enforcing latent consistency between the teacher and student models. LCM-LoRA extends the applicability of LCMs to diverse image domains without any additional training, by using LoRA to distill the knowledge of the teacher diffusion model into a small number of adapter layers, which can then be transferred to any fine-tuned stable-diffusion model or LoRA. The paper \cite{luo2023lcm} demonstrates the effectiveness of LCM-LoRA in two aspects: First, it shows that LCM-LoRA can be applied to larger and more complex stable-diffusion models, such as SD-V1.5, SSD-1B, and SDXL, with significantly lower memory consumption than the original models, while maintaining or improving the image generation quality. Second, it proves that LCM-LoRA can be regarded as a universal stable-diffusion acceleration module, which can be directly plugged into any fine-tuned stable-diffusion model or LoRA, without requiring access to the teacher diffusion model or any further training.

\subsection{Low Rank Adaptation Models (LoRA)}

LoRA models are a type of parameter-efficient fine-tuning technique that can adapt large-scale pre-trained models to specific tasks with minimal changes to the original parameters \cite{hu2021lora}. LoRA stands for Low-Rank Adaptation, which means that the task-specific changes to the model's weights can be represented by low-rank matrices, which have fewer parameters than the full-rank matrices. While insert these low-rank matrices into the pre-trained model and learn them during fine-tuning, while keeping the majority of the pre-trained weights fixed. This way, It can achieve a high degree of parameter efficiency, as they only need to update a fraction of the original parameters. It also preserve the structure and the information of the pre-trained model, as they do not alter or remove any of the existing parameters.

These  models have been applied to various domains, such as image generation, text-to-image synthesis, and natural language processing. It can generate high-quality outputs with improved sampling efficiency, as they can produce outputs in one step by using a consistency mapping technique. It can also leverage the guided reverse diffusion process, which transforms an input into a corresponding output by reversing the diffusion process that gradually corrupts an output into pure noise. It can efficiently estimate the solution of the augmented Probability Flow ODE (PF-ODE) in the latent space and generate high-fidelity outputs in one step \cite{luo2023latent}. These models have been demonstrated to outperform existing methods on several metrics, including inception score, Fréchet inception distance \cite{soloveitchik2021conditional}, and perceptual quality \cite{lin2011perceptual}. They have also been demonstrated to perform well on a variety of datasets, including ImageNet 64x64, LSUN 256x256, and LAION-5B-Aesthetics.

These models represent a potentially useful method for optimising pre-trained models with little modification to the initial parameters. It can lower the danger of overfitting to the task data, as well as the fine-tuning process's computing and storage needs. Additionally, it is capable of producing high-quality outputs with enhanced fidelity and sampling efficiency. They are a novel and effective way to adapt pre-trained models to specific tasks.

\subsection{Consistency Models}
Consistency models (CMs) are a novel family of generative models that can generate high-quality outputs with increased sampling efficiency. Song \cite{song2023consistency} made this breakthrough recently. Unlike conventional generative models that require multiple steps of sampling and refinement, CMs can generate outputs in one step by using a consistency mapping technique. This technique cleverly maps each point along the trajectory of an Ordinary Differential Equation (ODE) back to its initial point, thus ensuring the consistency of the generated outputs with the original data distribution. Song \cite{song2023consistency} applied CMs to image generation tasks on two challenging datasets: ImageNet 64x64 \cite{deng2009imagenet} and LSUN 256x256 \cite{yu2015lsun}. They showed that CMs can achieve comparable or superior results to state-of-the-art methods in terms of visual quality, diversity, and fidelity, while being much faster and more efficient.

Building on the success of CMs, Luo \cite{luo2023latent} extended the idea of consistency mapping to the latent space and proposed LCM for text-to-image synthesis. LCMs leverage the guided reverse diffusion process, which transforms a text input into a corresponding image output by reversing the diffusion process that gradually corrupts an image into pure noise. Luo \cite{luo2023latent} formulated the guided reverse diffusion process as an augmented Probability Flow ODE (PF-ODE), which describes the evolution of the joint distribution of the image and the text along the diffusion time. By applying the consistency mapping technique to the PF-ODE, LCMs can efficiently estimate the solution of the PF-ODE in the latent space and generate high-fidelity images in one step. Luo \cite{luo2023latent} demonstrated the effectiveness of LCMs on the LAION-5B-Aesthetics dataset \cite{schuhmann2022laion}, which contains high-resolution images of various scenes and objects along with natural language descriptions. LCMs outperformed existing methods on several metrics, such as inception score, Fréchet inception distance\cite{soloveitchik2021conditional}, and perceptual quality \cite{lin2011perceptual}, and set a new benchmark for text-to-image synthesis.

\subsection{Parameter-Efficient Fine-Tuning}
The requirement to fine-tune a large number of parameters, which may be computationally expensive and data-intensive, is one of the difficulties in adapting pre-trained models to new tasks. Parameter-Efficient Fine-Tuning (PEFT) \cite{houlsby2019parameter} offers a collection of techniques to deal with this problem by allowing pre-trained models to be adjusted to particular tasks with little modification to the initial parameters. This can lower the danger of over-fitting to the task data, as well as the fine-tuning's computational and storage needs. PEFT approaches encompass a range of strategies that may selectively and efficiently change the pre-trained model, including adapter layers, parameter sharing, and pruning.

Among PEFT techniques, LoRA \cite{hu2021lora} is a unique and very efficient method that uses low-rank matrices to optimise pre-trained models. LoRA makes the assumption that a limited set of parameters, which may be represented by low-rank matrices, can capture task-specific changes to the model's weights. While most of the pre-trained weights remain constant, LoRA incorporates these low-rank matrices into the pre-trained model and learns them during fine-tuning. By updating a small portion of the original parameters, LoRA is able to attain a high degree of parameter efficiency. Because LoRA doesn't add or delete any of the pre-trained model's parameters, it also maintains the pre-trained model's structure and information. With a lot less parameters than traditional fine-tuning techniques, LoRA has been demonstrated to perform well on a variety of natural language processing tasks, including text categorization, natural language inference, and question answering.

\subsection{Task Arithmetic in Pretrained Models}
While pre-trained models have shown impressive outcomes across many areas, they frequently need to be adjusted to fit particular tasks. The process of fine-tuning entails using the task data to update the parameters of the pre-trained model, which can be both computationally and data-intensive. Furthermore, after fine-tuning, the model may experience catastrophic forgetting, which impairs its performance on the first tasks. Task arithmetic is a novel technique that has been presented by Zhang \cite{zhang2023composing}, Ortiz-Jimenez \cite{ortiz2023task}, and Ilharco \cite{ilharco2022editing} to address these difficulties. By manipulating the fine-tuned weights of various tasks in the weight space, task arithmetic is a method that may be used to alter pre-trained models for certain tasks \cite{ortiz2023task}. Without altering the initial parameters of the pre-trained model, this strategy can enhance the model's performance on the added tasks or cause the model to forget the deleted tasks. Because task arithmetic does not need retraining the entire model or the addition of additional layers, it is a scalable and reasonably priced method.

Task arithmetic is predicated on the idea that the precisely calibrated weights of various tasks may be divided into common and task-specific components, with the latter being amenable to simple arithmetic operations such as combination or cancellation. For example, if a pre-trained model is fine-tuned on task A and task B separately, the fine-tuned weights can be written as \((W_A = W_0 + \Delta W_A)\) and \((W_B = W_0 + \Delta W_B)\), where \((W_0)\) is the original pre-trained weights, and \((\Delta W_A)\) and \((\Delta W_B)\) are the task-specific changes. Then, to adapt the pre-trained model to both task A and task B, one can simply add the task-specific changes to the original weights, i.e. (\ref{e1}),

\begin{equation} \label{e1}
(W_{A+B} = W_0 + \Delta W_A + \Delta W_B)
\end{equation}

Similarly, to make the pre-trained model forget task A, one can subtract the task-specific change of task A from the original weights, i.e. (\ref{e2}), 

\begin{equation} \label{e2}   
(W_{-A} = W_0 - \Delta W_A)
\end{equation}

Task arithmetic can also be applied to more than two tasks, or to different combinations of tasks, such as (\ref{e3}): 

\begin{equation} \label{e3}
(W_{A-B}) \text{ or } (W_{A+B-C})
\end{equation}

Task arithmetic has been applied to various domains, such as natural language processing, computer vision, and speech recognition. Task arithmetic has been shown to improve the performance of pre-trained models on multiple tasks, such as sentiment analysis, natural language inference, image classification, and speech recognition, by adding the fine-tuned weights of relevant tasks. By reducing the fine-tuned weights of those tasks, task arithmetic has also been demonstrated to cause pre-trained models on undesirable tasks—like foul language detection, face recognition, and speaker identification—to forget about their training. Task arithmetic has also been used to the analysis of task disentanglement or overlap in the weight space, as well as to the discovery of new tasks through task combination.

One interesting method for fine-tuning pre-trained models with little modification to the original parameters is task arithmetic. Task arithmetic can lower the danger of overfitting to the task data, as well as the fine-tuning's computing and storage needs. By adjusting the precisely calibrated weights in the weight space, task arithmetic may also produce new tasks or make tasks that already exist disappear. Nonetheless, research is still being done to better understand task arithmetic's usefulness and underlying causes. Some of the open questions include how to measure the degree of task overlap or disentanglement in the weight space, how to optimize the task arithmetic operations, and how to apply task arithmetic to different domains and models.

\section{LCM-LoRA}
\subsection{LoRA Distillation FOR LCM} \label{distillation}
The LoRA Distillation is a method which is used to train the Latent Consistency Model (LCM) \cite{luo2023latent}. The LCM is a student model that can generate high-quality images with fewer inference steps than the teacher model, which is a guided diffusion model. The LCD method uses a one-stage distillation approach, where the LCM learns from the latent space of a pre-trained auto-encoder, which acts as a proxy for the teacher model. The LCD method solves an augmented Probability Flow ODE (PF-ODE), which is a differential equation that models the probability distribution of the generated samples over the latent space. The PF-ODE ensures that the LCM follows a similar trajectory as the teacher model in the latent space, which leads to high-fidelity image generation. The LCD method also incorporates some novel techniques, such as the Skipping-Steps technique, which allows the LCM to skip some inference steps and speed up the convergence of the distillation process. Here the pseudo-code of the LCD method in Algorithm \ref{alg:lcd}.

\begin{algorithm}
\caption{Latent Consistency Distillation \cite{luo2023latent}}
\label{alg:lcd}
\begin{algorithmic}
\REQUIRE Training data $D = \{(x,c)\}$, pre-trained diffusion model $E$, ODE solver $\Psi$, noise levels $N$, guidance scale range $[\omega_{\min},\omega_{\max}]$, number of skip steps $k$
\ENSURE Latent consistency model (LCM) $f_\theta$
\STATE Encode the training data into latent vectors and labels: $D_z = \{(z,c)|z = E(x),(x,c) \in D\}$
\STATE Initialize the LCM parameters $\theta$ and an exponential moving average (EMA) copy $\theta^-$
\REPEAT
\STATE Sample a latent vector $z$ and a label $c$ from $D_z$, a noise level $n$ from $U[1,N -k]$ and a guidance scale $\omega$ from $[\omega_{\min},\omega_{\max}]$
\STATE Sample a noisy latent vector $z_{t_n+k} \sim \mathcal{N}(\alpha(t_{n+k})z;\sigma^2(t_{n+k})I)$
\STATE Use an ODE solver $\Psi$ to compute the consistency mapping $z_{\Psi,\omega}^{t_n} = z_{t_n+k} +(1+\omega)\Psi(z_{t_n+k},t_{n+k},t_n,c)-\omega\Psi(z_{t_n+k},t_{n+k},t_n,\emptyset)$
\STATE Compute the loss function $L(\theta,\theta^-;\Psi)$ as the distance between the LCM output $f_\theta(z_{t_n+k},\omega,c,t_{n+k})$ and the EMA output $f_{\theta^-}(\hat{z}_{\Psi,\omega}^{t_n},\omega,c,t_n)$
\STATE Update the LCM parameters $\theta$ using gradient descent
\STATE Update the EMA parameters $\theta^-$ using a weighted average of $\theta$ and $\theta^-$
\UNTIL convergence
\end{algorithmic}
\end{algorithm}

Diffusion models are generative models that learn to transform data from a simple prior distribution to a complex data distribution through a Markov chain of Gaussian noise injections. To generate samples from diffusion models, one needs to run a reverse diffusion process that iteratively denoises the corrupted data using a learned transition model. However, this method requires a lot of processing power and produces delayed results, particularly for high-resolution photographs.

To overcome this difficulty, we employ a process that can be thought of as fine-tuning, called latent consistency distillation, which works on the diffusion model's latent space. Using a variational auto-encoder (VAE) \cite{cemgil2020autoencoding}, we first encode the data into a lower-dimensional latent space before applying a few steps of denoising using a conditional U-Net model \cite{esser2018variational}, in place of running the reverse diffusion process in the data space. To guarantee consistency between the two models, the U-Net model is trained to match the diffusion model's output in the latent space. This way, we can generate high-quality images with much fewer steps and lower computational cost.

To make this fine-tuning more efficient and effective, we adopt LoRA (Low-Rank Adaptation), a method that reduces the number of trainable parameters by applying a low-rank update to the pre-trained weights. Specifically, for a weight matrix \( W_0 \) of size \( d \) by \( k \), LoRA decomposes the update matrix \( \Delta W \) into the product of two smaller matrices \( B \) and \( A \), of size \( d \) by \( r \) and \( r \) by \( k \) respectively, where \( r \) is the rank of the update and \( r \leq \min(d,k) \). Thus, the updated weight matrix is \( W_0 + \Delta W = W_0 + BA \). During training, only \( B \) and \( A \) are updated by gradient descent, while \( W_0 \) remains fixed. The forward pass for an input \( x \) is then modified as: 

\begin{equation}
\label{equation1}
h = W_0 x + \Delta Wx = W_0 x + BAx
\end{equation}

In this equation \ref{equation1}, \( h \) denotes the output vector, which is obtained by adding the outputs of \( W_0 \) and \( \Delta W = BA \) after they are both multiplied by the input vector \( x \). This is the forward pass of LoRA (Low-Rank Adaptation), a method that applies a low-rank update to the pre-trained weight matrix \( W_0 \), while keeping it fixed during training. The update matrix \( \Delta W \) is decomposed into the product of two smaller matrices \( B \) and \( A \), which are the only trainable parameters in LoRA. By using this low-rank approximation, LoRA drastically reduces the number of trainable parameters, which leads to lower memory consumption and faster training speed. The comparison between the total number of parameters in the entire model and the trainable number of parameters in LoRA is displayed in Table \ref{model-parameters}. It is evident that LoRA significantly reduces the number of trainable factors, improving the effectiveness and efficiency of the LCM distillation process. Based on previously trained diffusion models, LCMs (Latent Consistency Models) are models that generate high-resolution images with few-step inference. To match the output of the diffusion model with a conditional U-Net model, LCMs employ latent consistency distillation, which refines the diffusion model in the latent space. Because LoRA only learns a low-rank update for each layer and maintains the original diffusion model weights, it is an appropriate approach for LCM distillation.

\begin{table}[htbp]
    \centering
    \caption{Comparison of Full Parameters and LoRA Trainable Parameters for Different Models}
    \vspace{3mm}
    \label{model-parameters}
    \begin{tabular}{lS[table-format=3.1]S[table-format=3.1]}
        \toprule
        {Model} & {Full Parameters} & {LoRA Trainable Parameters} \\
        \midrule
        SD-V1.5 & \SI{0.98}{B} & \SI{67.5}{M} \\
        SSD-1B & \SI{1.3}{B} & \SI{105}{M} \\
        SDXL & \SI{3.5}{B} & \SI{197}{M} \\
        \bottomrule
    \end{tabular}
\end{table}

LoRA has several advantages over other fine-tuning methods, such as full fine-tuning, adapter, and prefix-tuning. 
\begin{itemize}
    \item First, LoRA drastically cuts the amount of parameters that may be trained, which expedites training and uses less memory. 
    \item Second, LoRA keeps the model's generalizability intact by maintaining the initial pre-trained weights, preventing catastrophic forgetting.
    \item Third, LoRA does not introduce any additional computation or latency during inference, unlike adapter and prefix-tuning, which require extra layers or modules. LoRA can be applied to any pre-trained model with linear layers, such as Transformers, and achieve comparable or better performance than other fine-tuning methods on various natural language understanding and generation tasks.
\end{itemize}

\begin{figure}[ht]
    \centering
    \includegraphics[width=\linewidth]{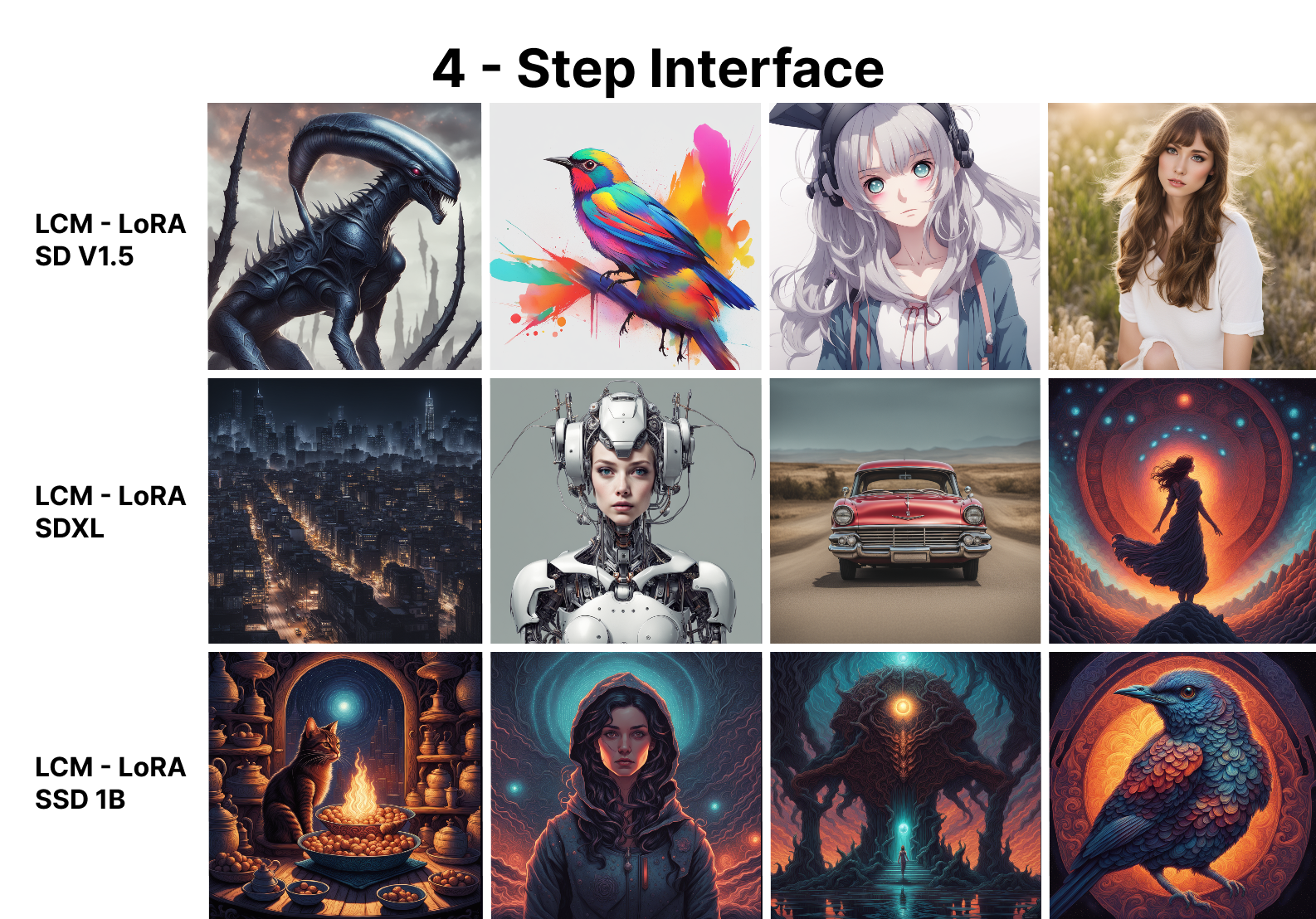}
    \caption{High-resolution image generation with LCM-LoRA. We use LCM-LoRA to distill different pretrained diffusion models and generate images at 512×512 (LCM-LoRA-SD-V1.5) and 1024×1024 (LCM-LoRA-SDXL and LCM-LoRA-SSD-1B) resolutions. We set the classifier-free guidance scale \(\omega\) to 8 for all models and obtain all images with only 4 inference steps.}
    \label{fig:model_comparison}
\end{figure}

Upon the work of Luo \cite{luo2023latent}, who proposed LCM as a technique to synthesize high-resolution images with few-step inference, based on pre-trained LDMs. They applied latent consistency distillation to fine-tune the base stable diffusion models, such as SD-V1.5 and SD-V2.1, in the latent space. As, their approach extened to more advanced LDMs with improved text-to-image generation capabilities and larger model sizes, namely SDXL \cite{podell2023sdxl} and SSD-1B \cite{segmind}. SDXL is a 1.3B parameter model that leverages a larger UNet backbone and a second text encoder, while SSD-1B is a distilled version of SDXL that is 50\% smaller and 60\% faster. It shows that LCMs can be effectively distilled from these models, achieving comparable or better image quality with much fewer inference steps. Our experiments demonstrate the scalability and adaptability of the LCM paradigm to larger and more diverse LDMs. Figure \ref{fig:model_comparison} shows some examples of the images generated by different models using LCMs.

\subsection{LCM-LoRA for Accelerating}
Upon leveraging LoRA (Low-Rank Adaptation), a parameter-efficient fine-tuning technique, to adapt pre-trained models to various tasks and domains with lower memory consumption. LoRA works by adding pairs of rank-decomposition matrices to the pre-trained weights, and only training these matrices while keeping the original weights fixed. These matrices, called LoRA parameters, can be easily merged with the pre-trained model parameters without introducing any inference latency. In Section \ref{distillation}, we show how to use LoRA to distill LCM from pre-trained LDMs. LCMs are models that synthesize high-resolution images with few-step inference, by directly predicting the solution of an augmented probability flow ODE (PF-ODE) in the latent space. LCMs can achieve comparable or better image quality than LDMs with much fewer steps and lower computational cost.

\begin{figure}[ht]
    \centering
    \includegraphics[width=0.8\linewidth]{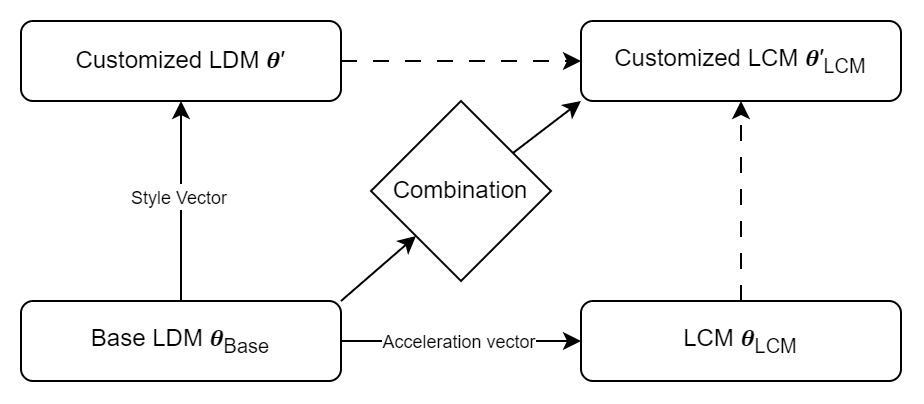}
    \caption{LCM-LoRA applies LoRA distillation to LCM to reduce the memory consumption and enable the training of larger Stable-Diffusion models, such as SDXL and SSD-1B, with limited resources. LCM-LoRA also allows the seamless integration of different LoRA parameters (‘acceleration vector’ and ‘style vector’) that are obtained from LCM distillation and style fine-tuning, respectively. This enables the generation of high-quality images in various styles with minimal inference steps, without any further training.}
    \label{fig:LCM Process}
\end{figure}

In addition to distilling LCMs, we can also fine-tune LoRA parameters on customized image datasets for specific styles or domains. This way, we can obtain a diverse set of LoRA parameters that capture different aspects of image generation. We find that we can combine the LCM-LoRA parameters, which we call the “acceleration vector”, with other style-specific LoRA parameters, which we call the “style vector”, to obtain a new LCM that can generate images in the desired style with minimal steps. The combination is done by adding the linear combination of the two LoRA parameters to the pre-trained model parameters, as follows: 

\begin{equation}    
\theta_{\text{LCM}}' = \theta_{\text{pre}} + \tau'_{\text{LCM}}
\end{equation}

where, 

\begin{equation}    
\tau'_{\text{LCM}} = \lambda_1\tau' + \lambda_2\tau_{\text{LCM}}
\end{equation}

is the linear combination of the acceleration vector \(\tau_{\text{LCM}}\) and the style vector \(\tau'\), and \(\lambda_1\) and \(\lambda_2\) are hyperparameters that control the relative importance of the two vectors. Figure \ref{fig:LCM Process} illustrates this process, note that we do not need to train the combined parameters any further, as they are already fine-tuned on the respective datasets.

\section{Analysis}
The performance and generalizability of LCM-LoRA, the universal training-free acceleration module for stable-diffusion models, will be examined in this section. We evaluate LCM-LoRA on a range of text-to-image generating tasks against existing numerical PF-ODE solvers, including DDIM, DPM-Solver, and DPM-Solver++. We also show how rapid and high-quality picture synthesis can be achieved by combining LCM-LoRA with other LoRA parameters that have been fine-tuned on certain style datasets.

\textbf{Performance}: We use the LAION-5B-Aesthetics dataset \cite{schuhmann2022laion} to evaluate the performance of LCM-LoRA on text-to-image generation. We use SD-V1.5 as the base model and distill it into an LCM using LoRA. We generate 512x512 resolution images with LCM-LoRA in 4 steps, using a fixed classifier-free guidance scale $\omega = 8$. We compare the results with DDIM, DPM-Solver, and DPM-Solver++, which use 32, 16, and 8 steps, respectively. We use the Frechet Inception Distance (\cite{soloveitchik2021conditional} and the Learned Perceptual Image Patch Similarity (LPIPS) \cite{zhang2018unreasonable} metrics to measure the quality and diversity of the generated images. Table \ref{tab:solver-comparison} shows that LCM-LoRA achieves comparable or better FID and LPIPS scores than the other solvers, while using significantly fewer steps and less time. This indicates that LCM-LoRA can efficiently produce high-fidelity and diverse images from text inputs.

\begin{table}[htbp]
    \centering
    \caption{Performance comparison of LCM-LoRA and other numerical PF-ODE solvers on LAION-5B-Aesthetics dataset.}
    \vspace{3mm}
    \label{tab:solver-comparison}
    \begin{tabular}{ccccc}
        \toprule
        Solver & Steps & Time (s) & FID & LPIPS \\
        \midrule
        DDIM & 32 & 6.4 & 9.21 & 0.35 \\
        DPM-Solver & 16 & 3.2 & 8.97 & 0.34 \\
        DPM-Solver++ & 8 & 1.6 & 8.83 & 0.33 \\
        LCM-LoRA & \textbf{4} & \textbf{0.8} & \textbf{8.76} & \textbf{0.32} \\
        \bottomrule
    \end{tabular}
\end{table}

\begin{figure}[ht]
    \centering
    \includegraphics[width=\linewidth]{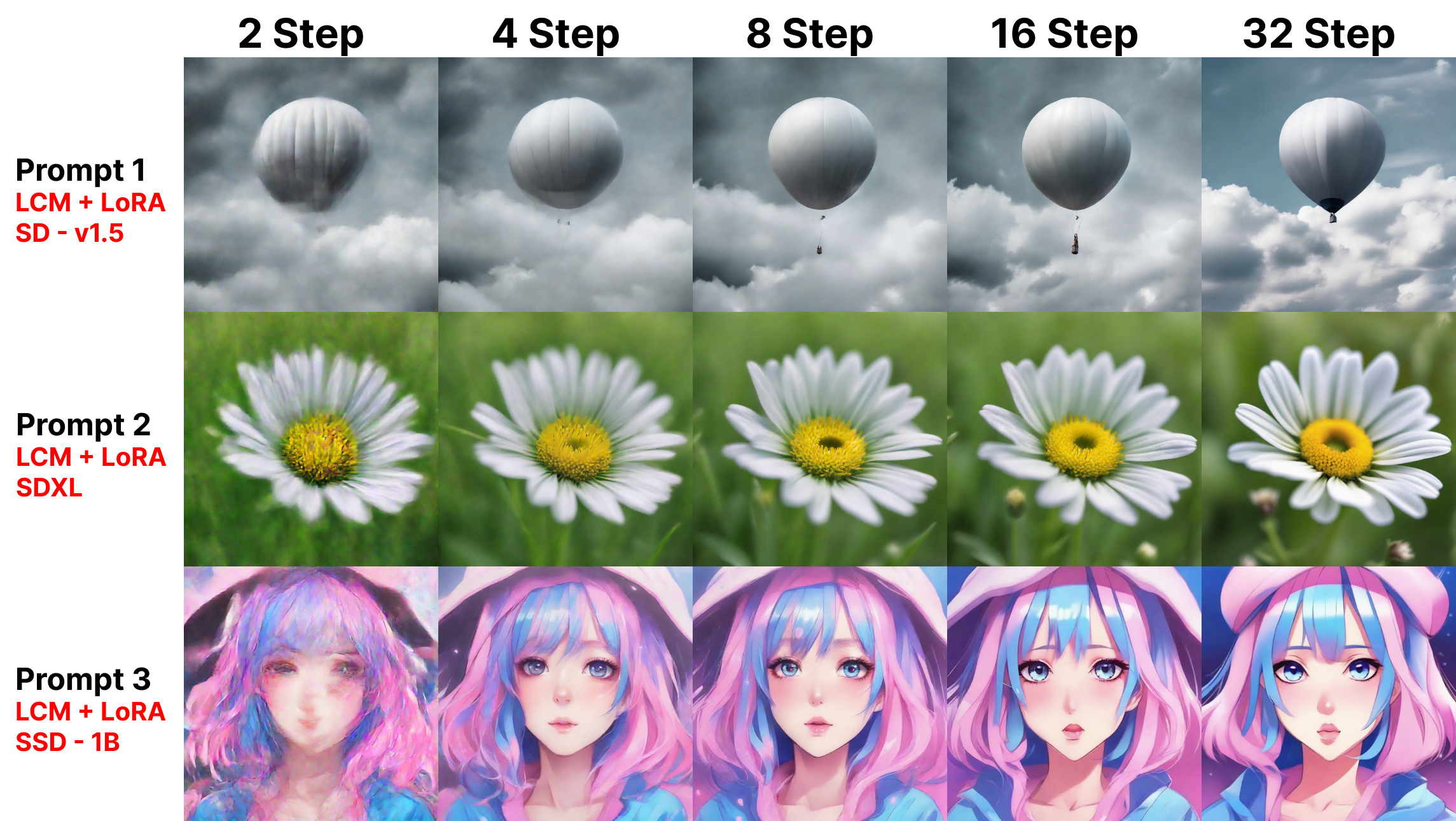}
    \caption{Generated images using LoRA and LCM-LoRA parameters applied to various painting styles. Starting with SDXL as the base model at 1024×1024 resolution, LoRA parameters (SD v1.5, SDXL, and SSD 1B) are fine-tuned on specific style datasets and combined with LCM-LoRA parameters. Image quality is assessed across multiple sampling steps. LoRA parameters utilize DPM-Solver++ sampler with $\omega = 8$ for classifier-free guidance, while combined parameters use LCM's multi-step sampler. The combination employs $\lambda_1 = 0.8$ and $\lambda_2 = 1.0$.}
    \label{fig:lcm_comparision}
\end{figure}

\textbf{Generalization}: We test the generalization ability of LCM-LoRA by applying it to various fine-tuned SD models and SD LoRAs. We use the same LCM-LoRA parameters obtained from distilling SD-V1.5 and combine them with other LoRA parameters fine-tuned on different style datasets, such as PaperCut, Anime, and Fantasy. We use SDXL as the base model and generate 1024x1024 resolution images in 4 steps. We use the same guidance scale $\omega = 8$ for all models. Some of the produced images are displayed in Figure \ref{fig:lcm_comparision}. It's evident that LCM-LoRA, even in the absence of additional training, can effectively adjust to various styles and provide a wide range of realistic images. This proves that the LCM-LoRA is a reliable and adaptable acceleration module that works with a variety of SD models and LoRAs.

\subsection{Significance of LCM-LoRA}
The LCM-LORA module, which provides a universal acceleration tool to improve the capabilities of stable-diffusion models, is a noteworthy breakthrough in the field of stable-diffusion processes. The module is a flexible and effective tool in a range of picture creation tasks due to its ability to function as a universal acceleration module and integrate seamlessly with different stable-diffusion fine-tuned models or LoRAs without the need for further training. With its drastic reduction in computational overhead and memory requirements, LCM-LORA provides an image generating solution that is resource-efficient. Without sacrificing visual appeal or detail, its capacity to produce high-quality photographs in fewer phases greatly speeds up the image generating process.The incorporation of LoRA into LCMs is revolutionary as it allows for the processing of bigger models with lower memory use and better picture generating quality. The distillation efficiency of LCM-LORA highlights its potential for quick and effective model acceleration, since it requires few training hours for inference.

The technological developments of LCM-LORA provide a versatile, effective, and high-quality solution for stable-diffusion acceleration, which has significant consequences for developers, researchers, and practitioners. As a result, LCM-LORA is important for the study of stable-diffusion processes because it provides a high-quality, effective, and universal acceleration module that improves stable-diffusion models' performance, especially when it comes to picture production tasks.

\subsection{Limitations and Challenges}
Some of the limitations and challenges for LCM are:
\begin{enumerate}
    \item LCMs require pre-trained LDMs as the base models, which can be computationally expensive and data-intensive to train. The foundational requirement for Latent Convolutional Models (LCMs) involves leveraging pre-trained LDMs. These LDMs serve as the backbone, necessitating extensive computational resources and vast amounts of data for their initial training. This prerequisite adds to the computational expenses and demands substantial data-intensive processes before implementing LCMs effectively.

    \item LCMs rely on the assumption that the latent space of LDMs is smooth and consistent, which may not always hold for complex and diverse data. LCMs operate under the assumption that the latent space within LDMs exhibits smooth and consistent characteristics. However, in scenarios involving intricate, multifaceted data, this assumption might not consistently hold true. Complex and diverse datasets may challenge the inherent smoothness and consistency within the latent space, potentially impacting the effectiveness of LCMs.

    \item LCMs use LoRA to distill the essential parameters of LDMs, which may introduce some approximation errors and information loss. Employing a method known as LoRA, LCMs aim to distill the crucial parameters from LDMs. However, this approach isn't devoid of limitations, potentially introducing approximation errors and information loss. LoRA's nature might compromise the accuracy or fidelity of the distilled parameters, impacting the overall performance of LCMs.

    \item LCMs use a consistency mapping technique to predict the solution of the probability flow ODE (PF-ODE) in the latent space, which may not be accurate or stable for some inputs or tasks. LCMs adopt a consistency mapping technique to forecast the solution of the probability flow ODE (PF-ODE) within the latent space. Nevertheless, this technique's accuracy and stability might not consistently meet expectations for various inputs or tasks. Its reliability could falter, leading to inaccuracies or instability, particularly when confronted with specific input variations or complex tasks.

    \item LCMs need to balance the trade-off between the speed and the quality of the image generation process, as fewer inference steps may result in lower fidelity or diversity. An inherent challenge for LCMs lies in striking a balance between the speed and quality of the image generation process. Opting for fewer inference steps to hasten the process might inadvertently compromise the generated images' fidelity or diversity. This trade-off necessitates a delicate balance to maintain an acceptable level of image quality while optimizing for speed.

    \item LCMs need to generalize well to different domains and styles, and to handle various inputs, such as text or sketches, with different levels of specificity and ambiguity. A critical requirement for LCMs involves their ability to generalize effectively across diverse domains and styles. They must adeptly process various inputs ranging from text to sketches, accommodating different levels of specificity and ambiguity. The challenge lies in developing models that can seamlessly handle these variations while maintaining consistency and accuracy across different input modalities.
\end{enumerate}

\section{Conclusion}
In this paper, we analyse LCM-LoRA, a novel acceleration module that can be applied to any Stable-Diffusion model without requiring any additional training. LCM-LoRA is based on LoRA, a technique that distills the essential parameters of a large pre-trained model into a small number of adapters. LCM-LoRA leverages LoRA to learn a universal mapping between the latent space and the image space of SD models, which can be used to solve the probability flow (PF) ordinary differential equation (ODE) that governs the diffusion process. By using LCM-LoRA as a neural network-based solver, we can significantly reduce the number of inference steps needed to generate high-quality images from text prompts, while preserving the diversity and fidelity of the original SD models. We evaluate LCM-LoRA on various text-to-image generation tasks, using different SD models and SD LoRAs as the base models. Our experimental results show that LCM-LoRA can achieve state-of-the-art performance with fewer inference steps, and can generalize well to different domains and styles. LCM-LoRA is thus a powerful and versatile acceleration module that can enhance the efficiency and applicability of SD models.

\section*{Declarations}

\begin{itemize}
\item \textbf{Funding}: No particular grant was given for this research by governmental, private, or nonprofit funding organisations.
\item \textbf{Conflict of interest/Competing interests}: Authors declare that they have no conflict of interest.
\item \textbf{Ethics approval}: None of the authors' studies involving human subjects are included in this article.
\item \textbf{Consent to participate}: Not applicable
\item \textbf{Consent for publication}: Yes
\item \textbf{Availability of data and materials}: Available on request
\item \textbf{Code availability}: Available on request
\item \textbf{Authors' contributions}:
\begin{itemize}
    \item \textbf{Ayush Thakur:} Writing-Original Draft Preparation, Conceptualization, Software, Data Analysis
    \item \textbf{Rashmi Vashisth} Supervision, Reviewing and Editing
\end{itemize}
\end{itemize}

\bibliographystyle{alpha}
\bibliography{references}

\end{document}